\documentclass{ifacconf}
\usepackage{comment}
\usepackage{tabularx}
\usepackage{booktabs}
\usepackage{subcaption}
\usepackage{caption} 
\usepackage{mathtools}
\usepackage{amsmath}    
\usepackage{amssymb}    
\usepackage{graphicx}   
\usepackage{bm}         
\usepackage{url}
\usepackage{graphicx}      
\usepackage{natbib}        
\usepackage{xcolor}
\usepackage[caption=false]{subfig} 
\usepackage{float}      
\begin{document}
\begin{frontmatter}

\title{Learning State-Space Models of Dynamic Systems from Arbitrary Data using Joint Embedding Predictive Architectures } 
\vspace{-0.2cm}
\thanks[footnoteinfo]{Authors have contributed equally}

\author[1]{Jonas Ulmen} 
\author[1]{Ganesh Sundaram} 
\author[1]{Daniel Görges}
\vspace{-0.2cm}
\address[1]{Department of Electrical and Computer Engineering, \\ RPTU University Kaiserslautern-Landau, Germany \\ 
    (e-mail: \{jonas.ulmen, ganesh.sundaram, daniel.goerges\}@rptu.de).}
\vspace{-0.2cm}

\begin{abstract}                
With the advent of Joint Embedding Predictive Architectures (JEPAs), which appear to be more capable than reconstruction-based methods, this paper introduces a novel technique for creating world models using continuous-time dynamic systems from arbitrary observation data. The proposed method integrates sequence embeddings with neural ordinary differential equations (neural ODEs). It employs loss functions that enforce contractive embeddings and Lipschitz constants in state transitions to construct a well-organized latent state space. The approach's effectiveness is demonstrated through the generation of structured latent state-space models for a simple pendulum system using only image data. This opens up a new technique for developing more general control algorithms and estimation techniques with broad applications in robotics.

\end{abstract}


\end{frontmatter}

\section{Introduction}
\vspace{-0.3cm}
State-space models and Markov decision processes are central concepts in control and reinforcement learning. The state variables capture the entire information of the system at any given time. Identifying this set is relatively straightforward for simpler systems. However, as the system's complexity increases, accurately determining the number of state variables required and distinguishing between state variables, parameters, and constants, and the relation between them becomes much more difficult. No reliable methods exist to construct state-space models for applications like robotic manipulation of soft objects, crops, fruits, cloth, or autonomous driving in complex and irregular environments. One promising method featuring neural networks in a Joint Embedding Predictive Architecture (JEPA) can learn informative latent variables without labeled data or by reconstructing the data itself. This study shows that JEPA can be adapted to span a latent state-space from video data and that making suitable predictions in the latent state space is possible. 


\vspace{-0.2cm}
\section{Related Work}
\label{sec:Related}
\vspace{-0.3cm}

Developing prediction models for environmental dynamics and control has been a long-standing area of research \citep{LJUNG20101, oh_action-conditional_2015, hafner_dream_2020, hafner_learning_2019, watter_embed_2015, shaj_action-conditional_2021, brunton2019}. Traditional approaches primarily focus on the construction of latent states that can reconstruct the input signal. However, such reconstruction-centric methods may not always produce informative representations for prediction tasks \citep{balestriero_learning_2024}.
To address this limitation, reconstruction-free contrastive approaches have gained significant attention \citep{okada_dreaming_2021, zhang_learning_2021, oord_representation_2019}. These methods leverage losses to learn embeddings that are broadly informative and transferable to downstream tasks. Contrastive methods \citep{caron_unsupervised_2020, chen_simple_2020, chen_exploring_2021, chen_intriguing_2021} typically involve corrupting input data to create augmented pairs, thereby preventing representational collapse \citep{jing_understanding_2022}. The model then learns invariance by linking the original input to its corrupted counterpart. Contrastive learning has been applied to latent space control \citep{shu_predictive_2020, deng_dreamerpro_2022, you_integrating_2022}.
Non-contrastive methods \citep{grill_bootstrap_2020, bardes_vicreg_2022, zbontar_barlow_2021} avoid explicit input corruption and instead rely on regularization techniques to ensure meaningful encodings. 
JEPA was proposed by \citet{lecun_path_2022}, adding a predictor network. An initial study used VICReg and InfoNCE losses to predict shifts of a dot under noise \citep{sobal_joint_2022}. A general pre-training approach for various vision tasks was proposed by \citet{assran_self-supervised_2023}. \citet{s_v_gradient-based_2023} proposed a latent space world model for planning and control, along with a control synthesis method. \citet{garrido_learning_2024} introduced a foundation model that learns a world model from images using JEPA. 
Control approaches that are quite similar to JEPA can be found in \cite{hansen_temporal_2022, hansen_td-mpc2_2024} and \cite{mondal_efficient_2023}. Here, actuated nonlinear systems are learned without reconstruction or contrastive methods, but in addition to state predictions, the rewards from the simulation environments are predicted as well, resulting in a latent state space that is partially conditioned on environment quantities that are not generally available in practical applications.
In \cite{sobal_joint_2022}, a simple RNN was used in the predictor module. When additional layers are added to an RNN and the prediction time steps shrink, the model approaches an ordinary differential equation in continuous time \citep{chen_neural_2019}.

\vspace{-0.2cm}
\section{Method}
\label{sec:method}
\vspace{-0.3cm}
\subsection{Contribution}
\vspace{-0.3cm}
We extend the approach in \citet{sobal_joint_2022} to adapt the architecture for dynamic tasks. We use a data sequence since a single measurement or image does not provide dynamic information, such as changes between instances. This sequence is stacked and fed into an observation encoder, which generates a continuous-time state vector for prediction. We also modify the predictor; Instead of having a feedforward network that maps the latent state from one discrete time step to the next, we employ the feedforward network as a model within an ODE integrator. Additionally, we introduce two new loss functions to ensure the smoothness of the state space. The first enforces a Lipschitz condition on the predictions over time, while the second, a contractive loss, ensures that inputs close to each other in the input space correspond to representations close to each other in the latent space. We demonstrate that the learned states are informative for both prediction and reconstruction.

\citet{sobal_joint_2022} introduced an example in which a dot was moved instantaneously from one time step to the next. With a wider perspective on dynamical systems, we select a simple pendulum actuated at its origin as our experimental setup. The transformations between each time instance involve movement induced by the actuation, which is integrated from acceleration to velocity and further to angles. We train a JEPA to model the dynamics of the pendulum in latent state space with an instantaneous state description that satisfies the Markov property.

\subsection{Architecture}
\begin{figure*}[h]
\centering
\includegraphics[width=0.9\textwidth]{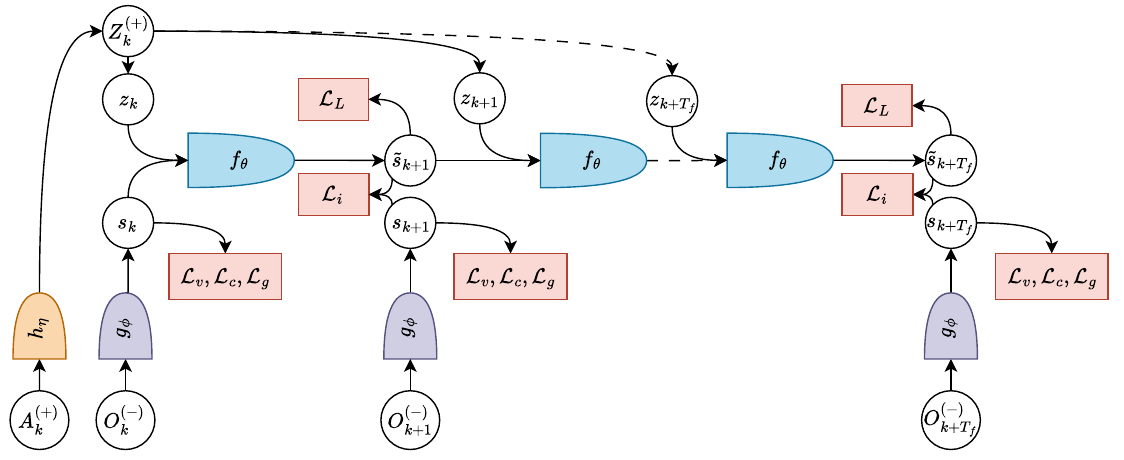}
\caption{Architecture with the flow of the training procedure for latent state-space construction and prediction described in Section \ref{sec:Training}. }

\label{fig:top_architecture}
\vspace{-0.1cm}
\end{figure*}

\vspace{-0.3cm}
We begin by considering a pre-recorded dataset of an observed Markov Decision Process (MDP) \( M = (O, A, P, R) \). Here, \( O \) represents the set of observations, \( A \) the set of actions, \( P = \Pr(o_k \mid o_{k-1}, a_{k-1}) \) denotes the transition probabilities, and the reward function is given by \( R: O \times A \rightarrow \mathbb{R} \). Our method aims to learn a prediction model within a latent space that functions as an actual state space, capturing the underlying dynamics of the MDP. The overall architecture is illustrated in Figure~\ref{fig:top_architecture}.

The dataset consists of observations \( o_k \in \mathbb{R}^{m_o} \) and their corresponding actions \( a_k \in \mathbb{R}^{m_a} \). The time step \mbox{\( k\in\mathbb{N}_0\)} indexes the sampling instants \( t_k \) obtained from the MDP by sampling with interval \(\Delta_t\) from the continuous time domain \( t\in\mathbb{R} \). 
To enable the prediction model to learn the dynamics of the system effectively, we use sequences of observations and actions. These sequences provide the temporal context necessary for understanding the evolution of the state.

\subsubsection{Encoders}
At each time step \( k \), we encode the sequence of previous observations \( O_k^{(-)} = [o_{k-T_p}, \dots, o_k] \), where \( T_p \) is the length of the past horizon. Using the past observation encoder 
\begin{equation}
g_\phi \big( O_k^{(-)} \big) = s_k, \quad
g_\phi: \mathbb{R}^{m_o \times T_p} \rightarrow \mathbb{R}^D,
\end{equation}
we map the observations into the \( D \)-dimensional latent state-space and obtain the latent state \( s_k \) for time step \( k \). The parameters of \( g \) are denoted by \( \phi \). The notation \( O_k^{(-)} \) indicates the ``past" sequence based on the time step \( k \). Similarly, the sequence of future observations is denoted as \( O_k^{(+)} = [o_{k+1}, \dots, o_{k + T_f}] \), where \( T_f \) represents the length of the future horizon. This notation will be used throughout the paper to describe sequences of any variable.

As for the observation sequences, the future action sequence \( A_k^{(+)} \) is encoded using the action encoder
\begin{equation}
h_\eta \big(A_k^{(+)} \big) = Z_k^{(+)}, \quad 
h_\eta: \mathbb{R}^{m_a \times (T_f-1)} \rightarrow \mathbb{R}^{D \times (T_f-1)},
\end{equation}
where each action is mapped into the \( D \)-dimensional latent state-space to obtain the future latent action sequence \( Z_k^{(+)} \), consisting of the latent actions \( z_k \).

During training, we also encode the sequences of past observations \( O_{k+1}^{(-)}, \dots, O_{k+T_f}^{(-)} \) for future time steps to obtain the sequence of future latent states \( S_{k+1}^{(+)} \).

\subsubsection{Predictor}
From the latent space, \( s_k \) and \( z_k \) are passed through a predictor
\begin{equation}
p_\theta (s_k, z_k) = \Tilde{s}_{k+1}, \quad
p_\theta: \mathbb{R}^{2D} \rightarrow \mathbb{R}^D,
\end{equation}
to predict the next latent state \( \Tilde{s}_{k+1} \), ideally achieving \( \Tilde{s}_{k+1} = s_{k+1} \) at the convergence of the parameters \( \theta \).
The predictor consists of two parts: Firstly, we employ a continuous-time neural ODE
\begin{equation}
\dot{s}_t = f_\theta(s_k, z_k), \quad
f_\theta: \mathbb{R}^{2D} \rightarrow \mathbb{R}^D,
\end{equation}
where \( \dot{s}_t \) is the time derivative of the state \( s_t \). Secondly, we use an integration method to simulate \( f_\theta(s_k, z_k) \) forward in time
\begin{equation}
s_{k+1} = l \big(f_\theta(s_k, z_k), s_k, z_k, \Delta_t \big), \quad
l: \mathbb{R}^{2D + D + D + 1} \rightarrow \mathbb{R}^D,
\end{equation}
where \( \Delta_t \) is the time step size for the simulation. Common integration schemes include the forward Euler or the 4-stage Runge-Kutta (RK4) method. In this paper, we use the latter. We perform the predictions auto-regressively until the future horizon \( T_f \) is reached, and therefore the prediction sequence of latent states \( \Tilde{S}_{k+1}^{(+)} \) is obtained.

\subsubsection{Observation Decoder}
To evaluate the performance of our method, we recover the observations \( \Tilde{O}_k \) from the latent state space into the observation space. After training the encoders and the predictor, we freeze their parameters and train an observation decoder
\begin{equation}
d_{\nu} \big(S_{k+1}^{(+)} \big) = \Tilde{O}_{k+1}^{(+)}, \quad
d_{\nu}: \mathbb{R}^{D \times (T_f-1)} \rightarrow \mathbb{R}^{O \times (T_f-1)}.
\end{equation}
The decoder has an inverse architecture to the encoder. Since our observations consist of images, we use a convolutional neural network for the observation encoder and transposed convolutions in the observation decoder. Note that \( S_{k+1}^{(+)} \), not \( \Tilde{S}_{k+1}^{(+)} \), is used for training. Otherwise, the decoders would be conditioned on the output of the predictor and encouraged to mitigate the flawed predictions of the predictor. During the evaluation, we use the latent state predictions \( d_{\nu}(\Tilde{S}_{k+1}^{(+)}) = \hat{O}_{k+1}^{(+)} \) to obtain the observation space representations of the predicted latent states.



\vspace{-0.3cm}
\subsection{Training}
\label{sec:Training}
\vspace{-0.3cm}
Training consists of two phases: First, we open a well-behaved latent state space by training the encoders and the predictor in a self-supervised manner. Secondly, the observation decoder is trained to recover observations from the latent state space for qualitative evaluation.

\subsubsection{Latent State Space Construction and Prediction}
\label{LatentStateSpaceConstructionandPrediction}
Let the state-space domain be defined by \mbox{\( S = \{s_k \mid k \in \mathbb{N}_0\} \)}. Let further \( \mathbf{s}_k \in \mathbb{R}^{N \times D} \) denote batched states and \( \mathbf{S}_k \in \mathbb{R}^{N \times T_f \times D} \) denote batched sequences, where \( N \) is the batch size. A single element is denoted by \( \mathbf{S}_k[n,\kappa,d] \) and \( \mathbf{S}_k[:,\kappa,d] \) stands for all elements along the first dimension.
The overall loss for this training phase is the weighted sum
\begin{equation}
\mathcal{L} = \lambda_1 \mathcal{L}_v + \lambda_2 \mathcal{L}_c + \lambda_3 \mathcal{L}_i + \lambda_4 \mathcal{L}_g + \lambda_5 \mathcal{L}_L,
\end{equation}
where \( \lambda_i \) are positive weighting factors. The main innovations to the self-supervised learning loss are the terms $\mathcal{L}_g$ and $\mathcal{L}_L$. We dissect the individual loss terms in the following.
The latent state space is spanned primarily using the VICReg (Variance-Invariance-Covariance-Regularization) loss functions \citep{bardes_vicreg_2022}.
The \textbf{Variance Loss} \(\mathcal{L}_v \) ensures the spread of the embeddings. The standard deviation for the \( d \)th state variable \( \mathbf{s}_{kd} \) is calculated as
\begin{equation}
\sigma(\mathbf{s}_{kd}) = \sqrt{\frac{1}{N-1} \sum_{n=1}^{N} 
\left( \mathbf{s}_{kd}[n] - \bar{\mathbf{s}}_{kd} \right)^2 + \epsilon_1 },
\end{equation}
where \( \epsilon_1 \) prevents numerical instabilities. The variance loss for the batched sequence \( \mathbf{S}_k^{(+)} \) is then
\begin{equation}
\mathcal{L}_v(\mathbf{S}_k^{(+)}, \epsilon_1, \epsilon_2) = 
\frac{1}{T_f D} 
\sum_{\kappa=1}^{T_f} \sum_{d=1}^{D} 
\frac{1}{\sigma(\mathbf{S}_k^{(+)}[:,\kappa,d], \epsilon_1) + \epsilon_2}.
\end{equation}

The \textbf{Invariance Loss} \(\mathcal{L}_i\) is defined as the mean squared simulation error, i.e.
\begin{align}
\mathcal{L}_{i}(\mathbf{S}_{k+1}^{(+)}, \mathbf{\Tilde{S}}_{k+1}^{(+)}) = 
& \frac{1}{(T_f-1) N} 
\sum_{\kappa=1}^{T_f-1} \sum_{n=1}^{N} \nonumber \\
& \left\| \mathbf{S}_{k+1}^{(+)}[\kappa,n] - \mathbf{\Tilde{S}}_{k+1}^{(+)}[\kappa,n] \right\|_2^2.
\end{align}

The \textbf{Covariance Loss} \( \mathcal{L}_c \) minimizes the off-diagonal elements of the covariance matrix to encourage unique embeddings, i.e.
\begin{equation}
\mathcal{L}_c(\mathbf{S}_k^{(+)}) = 
\frac{1}{T_f (N-1)} 
\sum_{\kappa=1}^{T_f} \sum_{i=1}^{D} \sum_{j=i+1}^{D} 
\left( 
\mathbf{S}_k^{(+)}[\kappa] \mathbf{S}_k^{(+)\top}[\kappa] 
\right)_{i,j}.
\end{equation}
To ensure a smooth state space, we newly introduce a \textbf{Contractive Loss} \( \mathcal{L}_g \) to the observation encoder. This loss relates to the variation between observations and embeddings, i.e.
\begin{align}
\mathcal{L}_g(\mathbf{S}_k^{(+)}, \mathbf{O}) = 
\frac{1}{T_f N} 
\sum_{\kappa=1}^{T_f} \sum_{n=1}^{N}
\left\| 
\frac{\partial \mathbf{S}_k^{(+)}[\kappa,n]}{\partial \mathbf{O}[\kappa,n]} 
\right\|^2_\mathrm{F},
\end{align}
where \( \| \cdot \|_F \) denotes the Frobenius norm.
To enforce smooth state transitions in the sense of a Lipschitz property, we furthermore newly apply a \textbf{Lipschitz Loss} \( \mathcal{L}_L \) to the predictor. This loss regularizes changes in predictor output, i.e.
\begin{align}
\mathcal{L}_L = 
& \frac{1}{(T_f-2) N D} 
\sum_{\kappa=1}^{T_f-2} \sum_{n=1}^{N} \sum_{d=1}^{D} \nonumber \\
& \max\left(0, \Delta p[\kappa,n] - L \Delta s[\kappa,n]\right)_d,
\end{align}

where
\begin{align}
\Delta p & = 
\big| p_\theta\big(\mathbf{S}_k^{(+)}[\kappa+1,n], \mathbf{Z}_k^{(+)}[\kappa+1,n]\big) \nonumber \\
& \quad - p_\theta\big(\mathbf{S}_k^{(+)}[\kappa,n], \mathbf{Z}_k^{(+)}[\kappa,n]\big) \big|, \\
\Delta s & = 
\big| \mathbf{S}_k^{(+)}[\kappa+1,n] - \mathbf{S}_k^{(+)}[\kappa,n] \big|.
\end{align}


\subsubsection{Observation Reconstruction}

To evaluate the predictions qualitatively, we train a decoder minimizing a reconstruction loss consisting of the Mean Squared Error (MSE) loss
\begin{align}
\mathcal{L}_{r,\text{mse}} = 
\frac{1}{(T_f-1) N} 
\sum_{\kappa=1}^{T_f-1} \sum_{n=1}^{N} 
\left\| \mathbf{O}_{k+1}^{(+)}[\kappa,n] - \mathbf{\Tilde{O}}_{k+1}^{(+)}[\kappa,n] \right\|_2^2
\end{align}
and the cosine similarity loss
\begin{align}
\mathcal{L}_{r,\text{cos}} = 
& \frac{1}{(T_f-1) N} 
\sum_{\kappa=1}^{T_f-1} \sum_{n=1}^{N} \nonumber \\
& \left( 1 - 
\frac{\mathbf{O}_{k+1}^{(+)}[\kappa,n] \cdot \mathbf{\Tilde{O}}_{k+1}^{(+)}[\kappa,n]}
{\|\mathbf{O}_{k+1}^{(+)}[\kappa,n]\|_2 \cdot \|\mathbf{\Tilde{O}}_{k+1}^{(+)}[\kappa,n]\|_2 + \epsilon} 
\right).
\end{align}

where \( \epsilon \) is a small scalar for numerical stability. The total reconstruction loss is
\begin{equation}
\mathcal{L}_r = \lambda_6 \mathcal{L}_{r,\text{mse}} + \lambda_7 \mathcal{L}_{r,\text{cos}}.
\end{equation}


\section{Implementation and Experiments}
\label{sec:result}
\vspace{-0.3cm}
For simplicity, we create a dataset from a simulated simple pendulum described by the differential equation

\begin{align*}
   \dot{x}(t) = 
\frac{d}{dt} \begin{pmatrix}
\theta(t) \\
\dot{\theta}(t)
\end{pmatrix} 
= 
\begin{pmatrix} 
    \dot{\theta}(t) \\ 
    -\frac{g}{L} \sin(\theta(t)) + \frac{1}{mL^2} \tau (t) \end{pmatrix}
\end{align*}

where \( g \) is the gravitational acceleration, \( L \)=2m is the pendulum length, $m$=2kg is the point mass, \( \theta \) is the pendulum angle, \( \tau \) is the input torque, and \( x(t) = [\theta(t),\dot{\theta}(t)]^T \) summarizes the state variables. We simulate the pendulum for $20{,}000$ steps with a sampling time of \( \Delta_t=0.1\,\mathrm{s} \). 
The pendulum is controlled using a PID controller, which generates a control signal \( \tau_{\mathrm{PID}}(t) \) based on the deviations between the current state $\theta(t)$ and the reference  $\theta^r(t)$. The gains of the controller are set to $K_p=500$, $K_i=0.2$, and $K_d=200$, which enable tracking the reference while maintaining nonlinear behavior. 
Reference values are repeatedly drawn from a uniform distribution over the admissible state space $[-\pi, \pi]$.

We capture the control signals as actions and represent the pendulum as grayscale images with \(64\times64 \) pixels, which serve as observations for the model. Both observations and actions are encoded into a latent dimension. Actions represented as 1D signals are processed using a three-block multilayer perceptron (MLP), where each block consists of a linear layer, dropout, and an exponential linear unit (ELU) activation. Each linear layer includes 128 neurons, chosen experimentally. Image observations are encoded using a small convolutional neural network with three blocks, each comprising a 2D convolutional layer, ELU activation, batch normalization, and dropout, followed by a linear layer and a sigmoid layer. For decoding, transposed convolutional layers of matching size are used, followed by a linear layer and a sigmoid layer.

During training, the predictor is fed a past sequence length of four steps (\(T_p = 4\)) for each observation (images) and the latest action. The prediction is a sequence of four future steps (\(T_f = 4\)) in the latent state space. The encoders and predictors are trained to convergence, followed by the decoder training until convergence. A latent dimension of \(D = 6\) was selected, which is ample compared to the pendulum's true state-space dimension~2.

In the first training phase, a parameter sweep was conducted for \(\lambda_1, \dots, \lambda_7\). The results showed that the covariance loss \(\mathcal{L}_c\) requires less emphasis than the variance loss \(\mathcal{L}_v\), consistent with the findings in \citep{bardes_vicreg_2022}. Even small weights for \(\mathcal{L}_c\) produced informative latent states. Analysis of the pendulum states revealed that the angle and angular velocity are not fully independent but vary distinctly. The contractive loss \(\mathcal{L}_g\) competes with the invariance loss \(\mathcal{L}_i\) when overly weighted. Reducing \(\mathcal{L}_g\) weight led to a greater reduction in \(\mathcal{L}_i\), highlighting a trade-off. The parameters satisfying \(\lambda_1 = \lambda_3 = \lambda_5 > \lambda_4 = \lambda_2\) yielded the best results.

\begin{figure*}[!th]
  \begin{subfigure}[t]{\textwidth}
    \centering
    \includegraphics[width=0.65\textwidth]{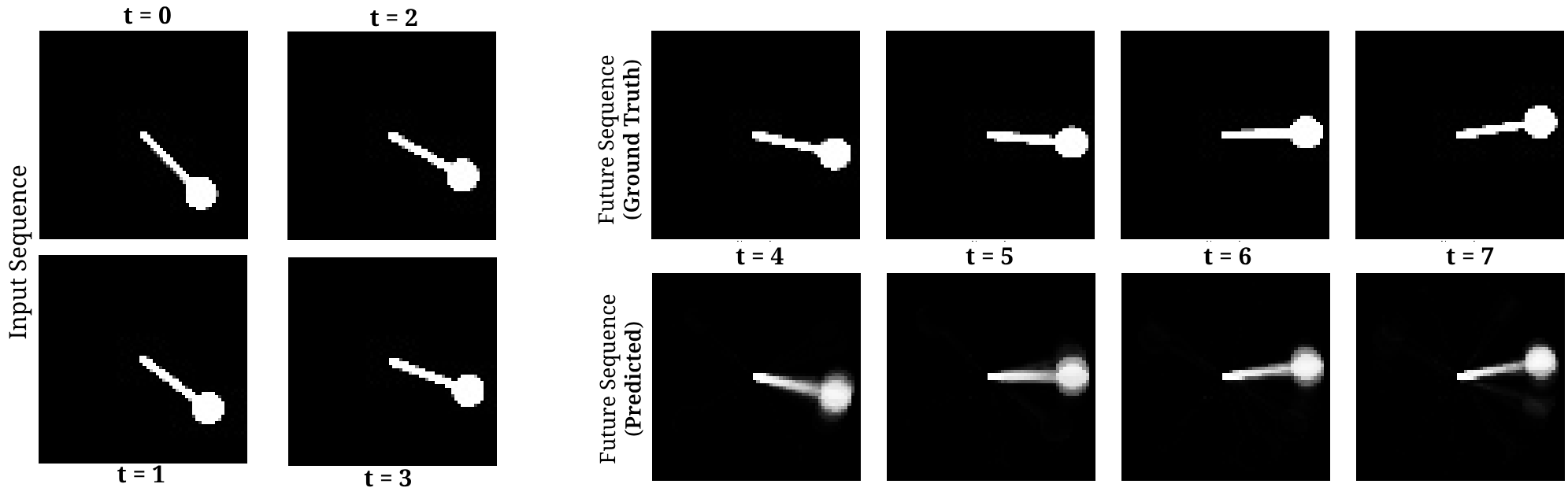}
    \vspace{-0.2cm}
    \caption{Experiment I}
  \end{subfigure}
  \vspace{-0.1cm}
  \begin{subfigure}[t]{\textwidth}
    \centering
    \includegraphics[width=0.65\textwidth]{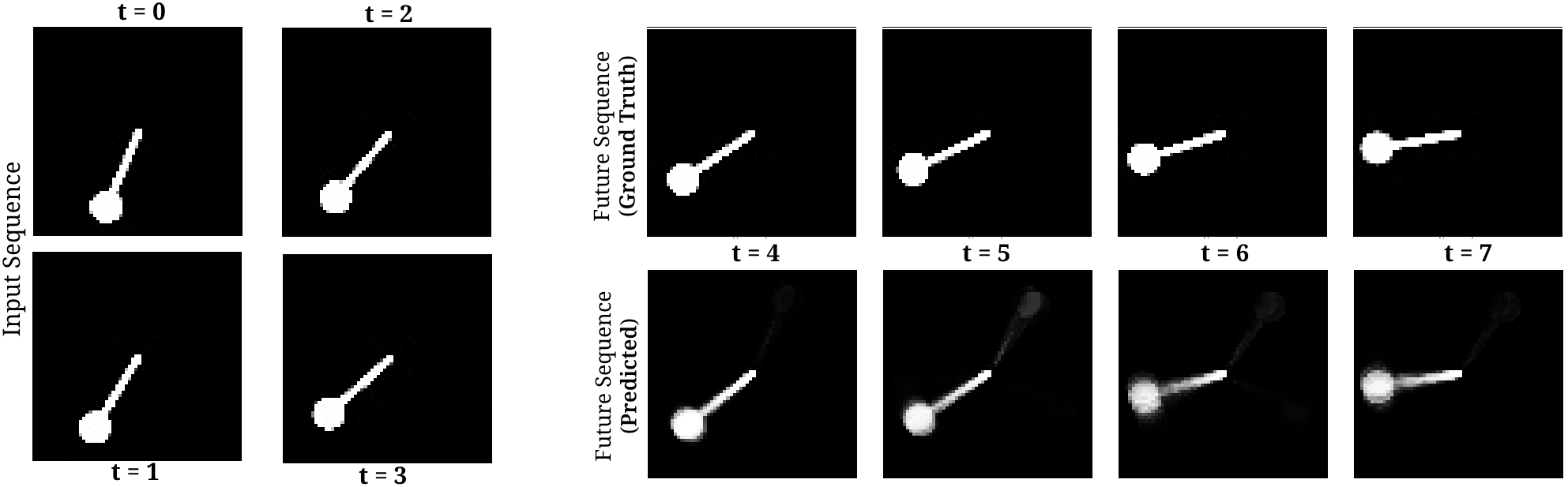}
    \vspace{-0.2cm}
    \caption{Experiment II}
  \end{subfigure}
  \begin{subfigure}[t]{\textwidth}
    \centering
    \includegraphics[width=0.65\textwidth]{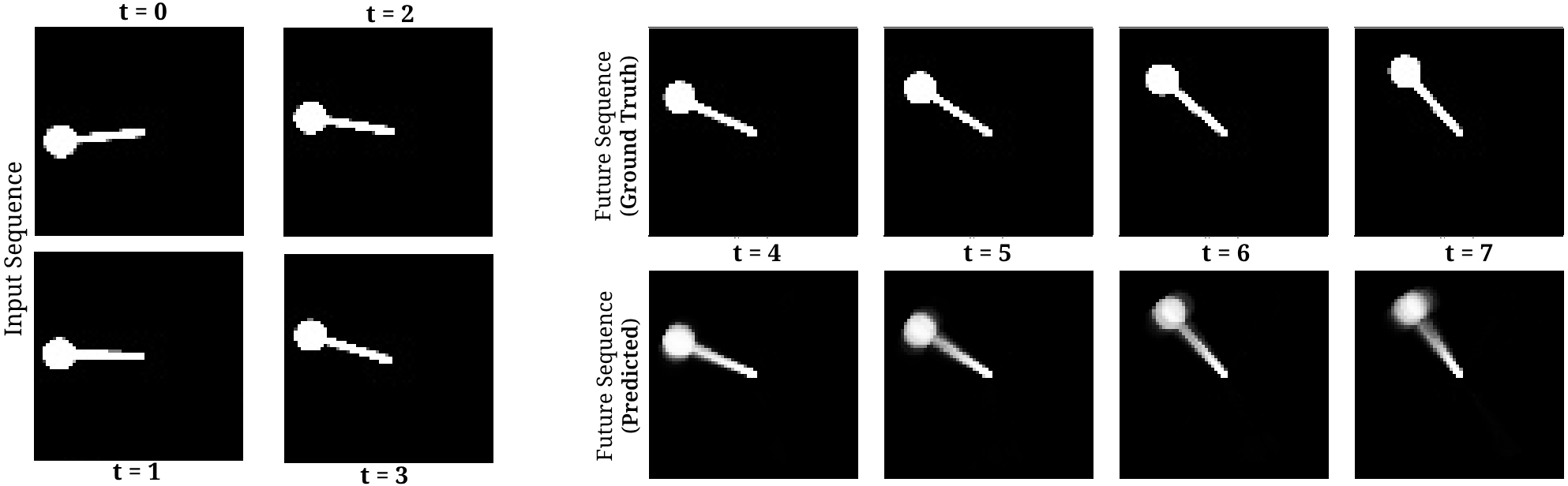}
    \vspace{-0.2cm}
    \caption{Experiment III}
  \end{subfigure}
  \vspace{-0.2cm}
  \caption{The figure presents the reconstruction results for qualitative analysis. It has to be noted that during this experimental evaluation, the losses defined in the study, $\mathcal{L}_L$ and $\mathcal{L}_g$, were utilized.}
\label{fig:combined_nice_results}
\end{figure*}

Qualitative validation through reconstructions demonstrated that the predicted pendulum angles closely match the ground truth images (Figure~\ref{fig:combined_nice_results}). The model effectively captured variations in angles, and changes in the latent state were accurately decoded into meaningful images.

\begin{figure*}[!th]
\centering
\begin{subfigure}[t]{\textwidth}
    \centering
    \includegraphics[width=0.65\textwidth]{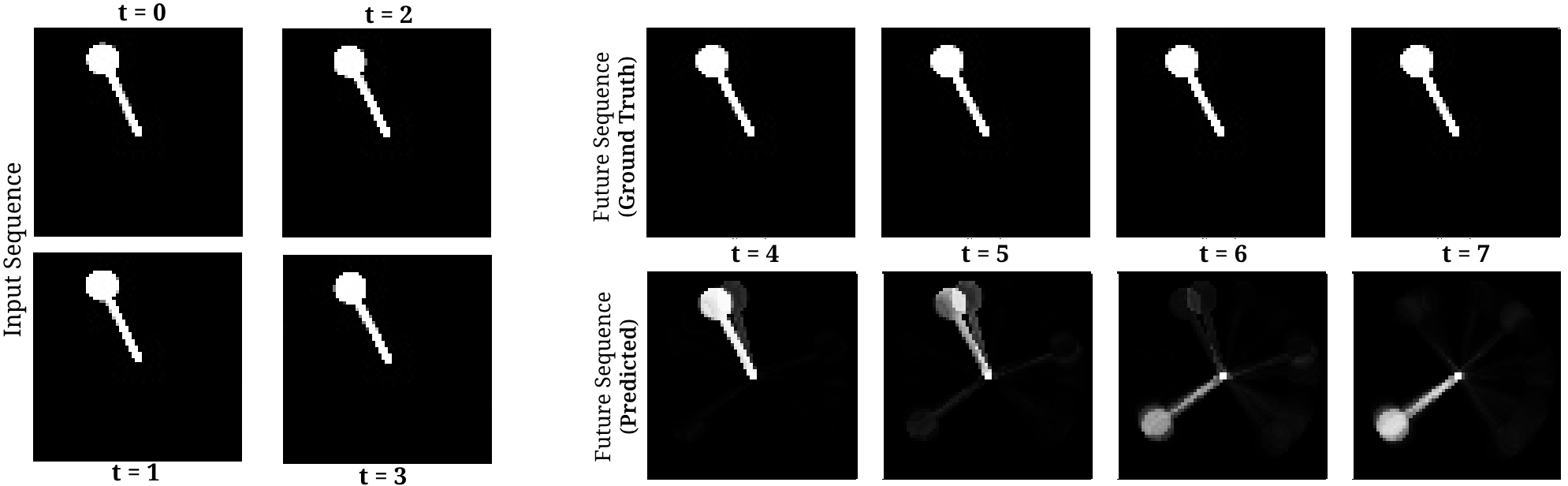}
    \label{fig:result_1b}
    \vspace{-0.2cm}
    \subcaption[]{Experiment I}
\end{subfigure}
\vspace{-0.2cm}
\begin{subfigure}[t]{\textwidth}
    \centering
    \includegraphics[width=0.65\textwidth]{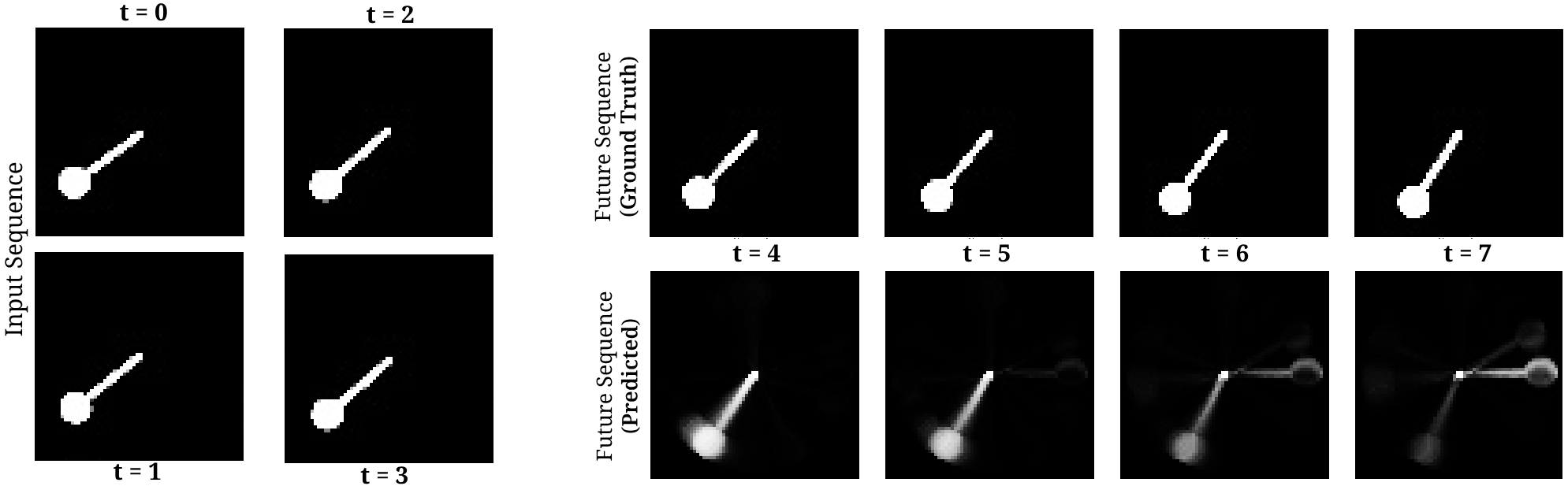}
    \label{fig:result_3b}
    \vspace{-0.2cm}
    \subcaption[]{Experiment II}
\end{subfigure}
\caption{The figure presents the reconstruction results for qualitative analysis. It has to be noted that during this experimental evaluation, the losses defined in the study, $\mathcal{L}_L$ and $\mathcal{L}_g$, were ignored.}
\label{fig:combined_bad_results}
\end{figure*}

The trials without incorporating \(\mathcal{L}_L\) and \(\mathcal{L}_g\) (Figure~\ref{fig:combined_bad_results}) showed noticeable shortcomings. Although the first two prediction steps were consistent, from the third step onward, the pendulum appeared blurry, and the angles exhibited significant jumps. This suggests that similar input images were not mapped to nearby latent states, amplifying minor prediction errors into substantial decoding discrepancies. Additionally, translucent instances of the pendulum were observed throughout the predicted images, likely caused by non-smooth and inconsistent mappings from observations to latent states. Incorporating \(\mathcal{L}_L\) and \(\mathcal{L}_g\) significantly improved the results but did not eliminate the translucent instances (Figure~\ref{fig:combined_nice_results}(b)).

We also experimented with different predictors. A simple feedforward neural network and a neural ODE with forward Euler integration performed poorly, though they worked on datasets with slow pendulum movements. However, beyond a certain actuation level, predictions became unreliable, and reconstructions degraded into simplistic outputs. The Runge-Kutta 4 (RK4) integration method provided reliable performance, suggesting that higher-order integration schemes could further enhance the predictor. Similarly, employing a more powerful encoder architecture could enrich the latent state space, easing the task for the predictor and improving overall performance.

We note that in some works, such as VICReg \citep{bardes_vicreg_2022} and JEPA \citep{lecun_path_2022}, feature expanders are used and certain losses are applied to the expander's output instead of applying them to the encoder's output. Since in our trials, we found that the incorporation of expanders did not improve our results and the main reference \citep{sobal_joint_2022} also did not incorporate expanders, we chose to omit them for brevity, simplicity, and conceptual comparability.

Reconstruction from a latent dimension of \(D=6\) was particularly challenging as it required the adaptation of codes that were found concerning another criterion. Unlike previous work \citep{sobal_joint_2022}, where the decoding into the image space was done with a larger latent dimension (\(D=512\)), we opted for the more complex task of reconstructing images directly from the lower-dimensional latent space.

\vspace{-0.2cm}
\section{Conclusion}
\label{sec:conclusion}
\vspace{-0.3cm}

We extended the JEPA approach to model dynamical systems by encoding sequences of image observations into a continuous-time latent state space. Using neural ODEs, the model enables predictions within this latent space. The approach was validated through the decoding of predicted states into image sequences of a pendulum system.

Although effective, several areas can be improved. The theoretical foundation for mapping sequenced data to continuous-time states needs strengthening. Additionally, the complexity of continuous-time state spaces necessitates more advanced integration schemes within the neural ODE. Image reconstruction from latent predictions remains challenging, as the encodings prioritize informativeness over reconstruction accuracy. The use of sophisticated integration methods and network architectures could address these challenges.

This study serves as an initial investigation. Further research is needed to establish this approach as a general framework for creating structured state spaces across diverse systems. Future work will include quantitative evaluations of various systems, handling different types of observation, and integrating state-of-the-art architectures like transformers. Incorporating theoretical properties like Lyapunov stability and contraction analysis will also be explored for system modeling and control.

\vspace{-0.2cm}
\bibliography{ifacconf}             


                                                   







\end{document}